\pdfoutput=1

\documentclass[11pt]{article}

\usepackage{acl}

\usepackage{times}
\usepackage{latexsym}
\usepackage{booktabs}
\usepackage{multirow}
\usepackage{float}

\usepackage[T1]{fontenc}

\usepackage[utf8]{inputenc}

\usepackage{microtype}
\usepackage{xspace}

\newcommand{\datasetname}[0]{Fig-QA\xspace}

\usepackage{xcolor}

\usepackage{hyperref}

\usepackage{graphicx}
\usepackage{tabularx}
\usepackage{footnote}

\usepackage{placeins}

\title{Testing the Ability of Language Models \\ to Interpret Figurative Language}

\author{Emmy Liu, Chenxuan Cui, Kenneth Zheng, Graham Neubig\\
  Language Technologies Institute \\
  Carnegie Mellon University \\
  \texttt{\{mengyan3,cxcui,kzheng2,gneubig\}@cs.cmu.edu} \\}

\begin{document}
\maketitle
\begin{abstract}
Figurative and metaphorical language are commonplace in discourse, and figurative expressions play an important role in communication and cognition. However, figurative language has been a relatively under-studied area in NLP, and it remains an open question to what extent modern language models can interpret nonliteral phrases. To address this question, we introduce \datasetname, a Winograd-style nonliteral language understanding task consisting of correctly interpreting paired figurative phrases with divergent meanings. We evaluate the performance of several state-of-the-art language models on this task, and find that although language models achieve performance significantly over chance, they still fall short of human performance, particularly in zero- or few-shot settings. This suggests that further work is needed to improve the nonliteral reasoning capabilities of language models.%
\footnote{Code and data are available at \url{https://github.com/nightingal3/Fig-QA}, and a leaderboard can be found at \url{https://explainaboard.inspiredco.ai/leaderboards?dataset=fig_qa}} 
\end{abstract}

\section{Introduction}




\textit{All our words are but crumbs that fall down from the feast of the mind} \cite{gibranpoem}. When humans read such a metaphorical phrase, how do they interpret it? Conceptual metaphors structure our everyday language and are used to map everyday physical experiences and emotions onto abstract concepts \cite{lakoff}. They allow us to communicate complex ideas, to emphasize emotions, and to make humorous statements \cite{fussel2008}. However, despite relating words in a way that differs from their accepted definition, these phrases are readily interpreted by human listeners, and are common in discourse \cite{Shutova2011ComputationalAT}, occurring on average every three sentences \cite{mio1996, fussel2008} 

The ability to interpret figurative language has been viewed as a bottleneck in natural language understanding, but it has not been studied as widely as literal language \cite{Shutova2011ComputationalAT,tong2021}.
Figurative language often relies on shared common-sense or cultural knowledge, and in some cases may be difficult to solve using language statistics. This presents a challenge to language models (LMs), as strong LMs trained only on text may not be able to make sense of the physical world, nor the social or cultural knowledge that language is grounded in \cite{bender-koller-2020-climbing, bisk-etal-2020-experience}.


Most previous work on figurative language focuses on metaphor detection, where a model is trained to \emph{identify} the existence of metaphors in text \cite{tsvetkov-etal-2014-metaphor, stowe-palmer-2018-leveraging, leong-etal-2020-report}, with datasets consisting mostly of conventionalized metaphors and idioms in wide use.
However, identifying these common metaphors that already appear often in language may be an easy task for LMs, and may not fully test their ability to interpret figurative language. The little work that exists on metaphor interpretation frames it as a task linking metaphorical phrases to literal rewordings, either through paraphrase detection \citep{bizzoni-lappin-2018-predicting} or paraphrase generation \citep{shutova-2010-automatic, SU2017300, mao-etal-2018-word} (details in \autoref{sec:related}) 
Another line of work probes for metaphorical understanding in LMs, but this is similar to the metaphor detection task, in that the LM is not actually asked to choose an interpretation for the metaphor \cite{pedinotti-etal-2021-howling, probing-metaphors}.
Although interesting, this work does not take into account the fact that metaphors are rich with different implications that may vary depending on the context. 

In this work, we ask whether or not LMs can correctly \emph{make inferences regarding creative, relatively novel metaphors} generated by humans.
This task is harder for two reasons: (1) \emph{inference} is harder than \emph{identification} or \emph{paraphrasing}, as it requires understanding the underlying semantics, and (2) the metaphors in our dataset are novel creations, and many may not appear even once in the LMs' training data.
We propose a minimal task inspired by the Winograd schema \cite{winograd}, where LMs are tasked with choosing the entailed phrase from two opposite metaphorical phrases. An example of a paired sentence is "Her commitment is as sturdy as (plywood/oak)". The correct answer would be either "She was (committed/uncommitted)". This can also be seen as an entailment task, where input $x$ is the premise, and the output $y$ is the hypothesis.%
\footnote{The opposing meanings help to avoid ambiguity in the correct answer, make the task intuitive for human annotators, and help prevent annotation artifacts that have plagued other NLI datasets \citep{gururangan-etal-2018-annotation}.}

\begin{table*}[!ht]
\centering
\small
\resizebox{0.85\textwidth}{!}{%
  \begin{tabular}{cc}
    \toprule \textbf{Paired sentences} & \textbf{Possible answers} \\ \toprule
    The pilot flew like a \underline{ballet dancer} & The pilot flew in a (\textbf{restrained way} | creative way)  \\
    The pilot flew like a \underline{modern dancer} & The pilot flew in a (restrained way | \textbf{creative way}) \\
    \midrule
    The meteor was as bright as \underline{New York City} & The meteor was (\textbf{very bright} | not bright at all)  \\
    The meteor was as bright as \underline{coal} & The meteor was (very bright | \textbf{not bright at all}) \\
    \midrule
    The atom is like a \underline{solar system} & Electrons (\textbf{orbit the nucleus} | are in probability densities)  \\
    The atom is like a \underline{cloud} & Electrons (orbit the nucleus | \textbf{are in probability densities}) \\
    \midrule
    He hustles like \underline{rent was due three days ago} & He (\textbf{hustles hardcore}. | doesn't hustle at all.)  \\
    He hustles like \underline{he's a billionaire's son.} & He (hustles hardcore | \textbf{doesn't hustle at all}) \\
    \midrule
    Life is as easy as \underline{kindergarten for a high school senior} & Life is (\textbf{basic} | beyond comprehension)  \\
    Life is as easy as \underline{kindergarten for a newborn} & Life is (basic | \textbf{beyond comprehension})  \\
    \bottomrule
  \end{tabular}}
  \caption{Example sentences from the dataset}
  \label{example-table} 
\end{table*}

We crowdsource a benchmark \textbf{\datasetname}, consisting of 10,256 such metaphors and implications (\autoref{sec:creation}), which can be used to evaluate the nonliteral reasoning abilities of LMs or for more broad studies of figurative language in general (we provide preliminary analyses in \autoref{sec:typologies}).
Through extensive experiments on strong pre-trained LMs (\autoref{sec:baseline-evaluation}), we find that although they can be fine-tuned to do reasonably well, their few-shot performance falls significantly short of human performance (\autoref{sec:results}).
An in-depth analysis (\autoref{sec:error-analysis}) uncovers several insights: (1) LMs do not make use of the metaphorical context well, instead relying on the predicted probability of interpretations alone, (2) the task of associating a metaphor with an interpretation is more difficult than the reverse, (3) even strong models such as GPT-3 make inexplicable errors that are not well-aligned with human ones, indicating that further work is needed to properly model nonliteral language.

\section{Dataset Creation and Validation}
\label{sec:creation}

\subsection{Crowdsourcing Task}
We crowdsourced data from workers on Amazon Mechanical Turk ( details in \autoref{sec:crowdsourcing}). Workers were asked to generate paired metaphors with different meanings, as well as literal implications of the two metaphors in context. We instructed workers to try to generate rare or creative metaphors, namely ``metaphors that would not appear often in text on the internet, books, social media, or news sites, but that can still be easily understood by people.''
Workers were given examples of valid pairs that fit the format, and examples of invalid ones to discourage errors. 
Some examples of generated pairs are shown in \autoref{example-table}.

To help workers, we employ the \textit{randomness as genesis} and \textit{narrow limits of change} principles of Cognitive Load Theory \citep{cognitiveload}. 
To add soft constraints, we generate 3 different random words to be shown to each batch of workers. However, workers were not required to use these words, as we wanted to encourage maximum diversity. To ensure that the random words were metaphorically rich, we selected them based on the metaphorical frames in \citet{lakoff}. 

\subsection{Data Validation}
The dataset was manually validated by three authors of this article. Each author covered roughly one-third, evenly split between training, validation, and test. Examples were excluded if they (a) did not make sense given the figurative expression, (b) had grammar or spelling errors that rendered them unintelligible, or (c) did not follow the format of the task. Examples of excluded samples are included in \autoref{sec:invalid}.
We collected 13,324 sentences and interpretations from the crowdsourcing task, and 10,256 sentences remained after filtering.

\subsection{Final Dataset}

The release version of our dataset contains the named data splits in \autoref{data-splits}. The medium train, dev, and test splits were generated by partitioning the first stage of the data collected. The large train split additionally contains all the new examples from the second collection stage, and the small train split is a small random sample.

\begin{table}[!ht]
    \centering
    \small
    \begin{tabular}{ccc|c|c}
        \toprule
        \multicolumn{3}{c|}{Train} & \multirow{2}{*}{Dev} & \multirow{2}{*}{Test}  \\
        S & M & L &  &  \\ \midrule
        200 & 1,458 & 8,016 & 1,094 & 1,146 \\
        \bottomrule
    \end{tabular}
    \caption{Examples in each data split}
    \label{data-splits}
\end{table}


\section{Figurative Language Typologies}
\label{sec:typologies}
In this sample, we perform an analysis of the collected data to demonstrate its trends and categorize examples for further error analysis.
Specifically, we examine (a) subjects, objects, and relations, and (b) types of common-sense knowledge needed to interpret the metaphor.

\subsection{Figurative Language Structure}
\label{sec:sro}
\begin{figure}[!t]
    \centering
    \includegraphics[width=0.85\columnwidth]{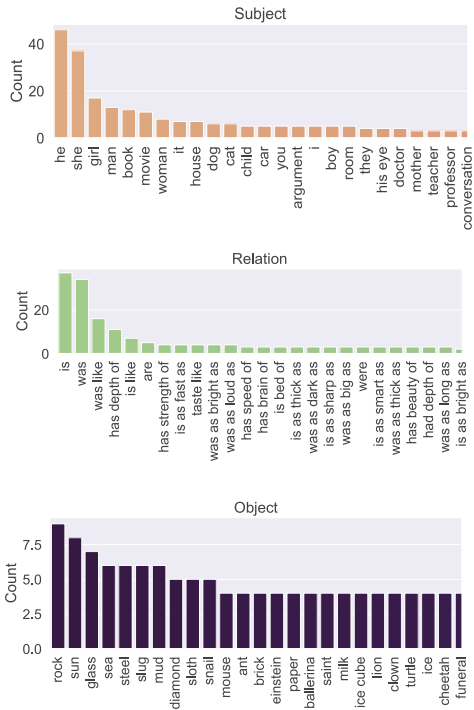}
    \caption{Visualization of 25 most frequent subjects, relations, and objects in the medium train set.}
    \label{metaphor-triplet}
\end{figure}

We note that most metaphors and similes can be characterized by three components, $(S, R, O)$, where $S$ is a subject, $R$ is a relation, and $O$ is an object. For instance, "Her commitment is as sturdy as plywood" can be written (Her commitment, sturdy, plywood). Interpretation involves inferring an attribute of the subject by extracting a relational attribute from the object \cite{Fauconnier2003ConceptualBF}. In a simile, $R$ is explicit, whereas it is usually implicit in a metaphor. The most common subjects, relations, and objects in the medium train dataset are shown in \autoref{metaphor-triplet}. These were obtained by first segmenting the phrases with syntactic patterns constructed from observation, followed by lemmatization and removal of punctuation and determiners "the", "an", "a" and "that". 
There are 441 unique subjects, 646 unique relations, and 1,198 unique objects in the medium training set.

\subsection{Common-sense Knowledge Types}
\label{sec:metaphor-typology}
\begin{table*}[t]
\centering
\resizebox{0.85\textwidth}{!}{%
  \begin{tabular}{cc}
    \toprule 
    \textbf{Type of knowledge required} &\textbf{Paired sentences} \\ \toprule
    Common-sense (objects) & The new mattress is just as comfortable as sleeping on a (cloud/rocks outside) \\ \midrule
    Visual & The professor's argument had the clarity of a (crystal glass/marine fog) \\ \midrule
    Common-sense (social) & She is as embarrassed as a kid that (forgot homework/got an A) \\ \midrule
    Cultural & The construction was as disastrous as the (1981 musical Cats/The 2019 film based on the musical Cats) \\ \bottomrule
  \end{tabular}}
  \caption{\label{metaphor-types} Metaphor types based on types of knowledge required (not mutually exclusive)}
\end{table*}

Next, we examined the test set to determine the types of commonsense knowledge needed to interpret metaphors. Through thematic analysis, we devised 4 categories based on common-sense knowledge, which are not mutually exclusive: common-sense object knowledge, visual metaphors, common-sense social understanding, and cultural knowledge. The same 3 authors of the paper annotated the test set for these categories, with annotators responsible for separate categories.

\textbf{Common-sense object knowledge} consisted of metaphors that made reference to properties of common objects and animals, such as volume, height or mass of objects, or properties of materials.  68.35\% of the test set was found to require common-sense object knowledge.

\textbf{Visual metaphors} were a subset of common-sense object metaphors, based primarily on the visual modality, including attributes such as brightness or colour. Some visual metaphors also sketched a vivid visual scene. These examples comprised 14.73\% of the test set.

\textbf{Common-sense social understanding} examples required knowing how humans would react in different circumstances or required knowing about human emotions. These examples comprised 27.55\% of the test set.

\textbf{Cultural metaphors} required knowing cultural traditions, works of art/artefacts, or religion. Due to crowdworkers being recruited from the US, these are centered around US culture. These examples comprised 16.56\% of the test set.


\section{Baseline Models and Evaluation}
\label{sec:baseline-evaluation}
\subsection{Auto-regressive Language Models}
Auto-regressive LMs generate a probability distribution of the next token given all preceding tokens. As such, we can directly compute the probability of a sentence by multiplying the conditional probability of each token at each time step. 
$$ \tilde{P}(w_1...w_N) = p(w_1) \prod_{i=2}^{N}{p(w_i|w_1...w_{i-1})}$$

The ability to directly extract probabilities enables the \textit{zero-shot} reasoning of these LMs. For a pair of metaphorical expressions $x_1$ and $x_2$ with two corresponding interpretations $y_1$ and $y_2$, we feed in the concatenation of the metaphor and the interpretation to the pretrained model without finetuning.
We define ``forward'' and ``backward'' probabilities assigned to interpretations and figurative language expressions, respectively.
For the \textbf{forward probability}, for figurative phrase $x_i$ and correct answer $y_i$, we take $$P(y_i|x_i) = \frac{P(x_i,y_i)}{P(x_i,y_i) + P(x_i,y_j)}$$ since there are only two answer options. From this, we can calculate accuracy when we take the indicator of $P(y_i|x_i) > 0.5$. Similarly for the \textbf{backward probability} (predicting phrase based on answer), we take $$P(x_i|y_i) = \frac{P(x_i, y_i)}{P(x_i, y_i) + P(x_j, y_i)}$$
with analogous backward accuracy.%
\footnote{
In actuality, we use the length-normalized probability that a model assigns to a sentence as a heuristic for the total probability, to minimize the effect that the length of a sentence has on the decision (though this is not the probability of the sequence in a strict sense): $P(w_1...w_N) = \exp(-\frac{1}{N} \log \tilde{P}(w_1...w_N))$.
Initial experimentation showed marginal differences in accuracy when using these two methods, so we used normalized probabilities by default.
}

We examine three state-of-the-art large transformer-based LMs of this category: \textbf{GPT-2} (with 117M parameters, trained on 40GB of text), \textbf{GPT-neo} (with 1.3B parameters, trained on 800GB of text) and \textbf{GPT-3} (4 variants between 350M and 175B parameters, trained on 45TB on text) \cite{radford2019language, gpt-neo, gpt3}.
We also examine the performance of these models after finetuning on the training data. 
GPT-2 and GPT-neo were trained with a batch size of 8, with early stopping with patience of 1 epoch, and a minimal hyperparameter search was performed with learning rates 1e-5 to 5e-5.
GPT-3 was trained with the default parameters of the GPT-3 finetuning API.

\subsection{Masked Language Models}
We also evaluate the performance of masked LMs on this task. Unlike auto-regressive LMs, masked LMs cannot directly output the probability of a sentence, so it is not possible to directly test the zero-shot performance of these models. Instead, we test the transfer performance by first finetuning them in two ways: on WinoGrande, which is also a binary choice task based on common-sense reasoning, and on several NLI datasets, including SNLI, MNLI, FEVER-NLI and ANLI \cite{nli-model, Sakaguchi2020WINOGRANDEAA}. The input to the model trained on WINOGRANDE is formatted as \verb|[CLS][metaphor][SEP]| \verb|[answer1][SEP][answer2]|, and we use an extra linear layer on the \verb|[CLS]| token embedding to perform the classification. In addition to the transfer performance, we also use contrastive fine-tuning by feeding in each metaphor along with both answer choices, and training the model with our dataset to classify which answer is correct. 
For the NLI model, we examine accuracy using all three labels the model was originally trained with (entailment, neutral, and contradiction), as well as using a forced binary choice paradigm in which the logits for the contradiction label are subtracted from the logits for the entailment label, and the higher "entailment score" is the ending the model predicts.
We treat these two conditions as the analog of ``zero-shot" for these models.

We examine two masked LMs that are commonly used as baselines on many NLP tasks: \textbf{BERT} \citep{devlin2019bert}, a transformer-based LM jointly trained on the masked LM and next-sentence prediction objectives, and \textbf{RoBERTa} \citep{liu2019roberta}, an improved variant of BERT 
which consistently outperforms BERT across most tasks. We use the large variant of both models (350M parameters). BERT and RoBERTa were finetuned on the medium dataset for 8 epochs with batch size 8, following the setting in \cite{Sakaguchi2020WINOGRANDEAA}. A hyperparameter search was done with learning rates 5e-6 to 2e-5. Both BERT and RoBERTa were used for the Winogrande experiments, while only RoBERTa was used for the NLI experiment.

\subsection{Forced-choice Paradigm}

Due to the inherent creativity of metaphors, there may be different interpretations of the same metaphor. For instance, in \autoref{example-table}, the example "he hustles like he's a billionaire's son" could also be interpreted in other ways, for instance "he uses his father's contacts and social privileges to make money". In a structural-mapping context, the forced choice between two answers constrains the possible meaning of the metaphor to be along one axis \cite{structure-mapping}. In this case, it would be whether or not he is required to work hard.

Of course, many of these metaphors have other valid interpretations. In the "billionaire's son" example, another valid axis of interpretation could be the manner in which he works. For instance, the alternative pair could be "he hustles like he's a (billionaire's son | single mother working three jobs)" with answers "he (uses his contacts and social privileges to make money | works extremely long hours with multiple ventures to make money)". It is possible that LMs could come up with other valid interpretations that are not the ones originally intended, motivating us to also look at generation performance in section \autoref{sec:generation-res}.


\subsection{Human Performance}


To estimate the expected human performance on this task, we ran a benchmark on the test set with 10 human volunteers who were not the authors of the article. The human annotators were not shown any training examples, so this would be equivalent to the zero-shot setting for models. Participants ranged from 20 to 29 years old, and there were 5 male and 5 female participants. 5 each were native- and non-native English speakers respectively. Participants were mainly graduate student volunteers.

We shuffled the test set and split it into 10 partitions of $\approx$115 examples for each annotator. The examples were presented with pairs shuffled and separated, in order to create a better comparison with model performance.

Due to differences in vocabulary or cultural background, we instructed participants to mark examples where they weren't confident, such as those that contained words or cultural references they didn't understand.

\section{Results}
\label{sec:results}



\subsection{Inference Results}

The first question is \textbf{whether strong LMs can interpret metaphors at all when presented with two opposing meanings, in zero-shot or supervised settings}. 
These results are presented in \autoref{results-table}. The results for masked language models are higher than those for autoregressive language models, and fine-tuning significantly improves performance for all models. 


\begin{savenotes}
\begin{table}[!t]
    \centering
    \resizebox{\columnwidth}{!}{ 
    \begin{tabular}{cccc}
        \toprule
        Model & Zero-shot & Tuned (L) & Tuned (XL) \\ \toprule
        
        GPT-2           & 53.93             & 54.80 & 62.65 \\
        GPT-neo 1.3B    & 56.89             & 69.98 & 72.00\\
        GPT-3 Ada       & 59.08             & 69.17 & 73.56\\
        GPT-3 Babbage   & 62.91             & 73.97 & 77.31\\
        GPT-3 Curie     & 65.35             & \textbf{79.04} & \textbf{81.94} \\
        GPT-3 Davinci   & \textbf{68.41}    & - & - \\ \midrule
        BERT            & 58.14             & 83.16 & 85.69 \\
        RoBERTa         & \textbf{66.18\footnote{This is the accuracy score when transferring from Winogrande. Pretrained NLI results were 50.47 when using original labels (entailment/contradiction/neutral), and 66.32 when forcing a binary decision.}}             & \textbf{89.22} & \textbf{90.32} \\ 
        \midrule
        Human & 94.42 & - & -\\
        Human (confident) & \textbf{95.39} & - & - \\
        \bottomrule
    \end{tabular}
    }
    \caption{Zero-shot and finetuned test accuracies (\%), finetuned is averaged across 5 seeds. Dev set accuracies can be found on the \href{https://explainaboard.inspiredco.ai/leaderboards?dataset=fig_qa}{leaderboard} under the "validation" split.}
    \label{results-table}
\end{table}
\end{savenotes}


\paragraph{Zero-shot Performance}
For the zero-shot setting, we examine the test accuracy based on zero-shot forward probabilities for the GPT models, and the pseudo "zero-shot" transfer performance for BERT and RoBERTa using models pretrained on the WinoGrande task \cite{Sakaguchi2020WINOGRANDEAA}. As shown, the GPT-3 models outperform the GPT-2 and GPT-neo models. Among the GPT-3 models, there is a clear correlation between model size and performance, with the largest model (GPT-3 Davinci) achieving the highest zero-shot test accuracy. BERT and RoBERTa achieve accuracy within the range of GPT-3 models. While our models mostly perform much better than chance in the zero-shot setting, there is still a large gap of 26 percentage points between our best model and human level performance.

\paragraph{Fine-tuned Performance}
For the fine-tuned setting, all models listed are fine-tuned on the small data set split. GPT models were trained with language modeling loss, whereas BERT and RoBERTa are trained with contrastive loss. We did not evaluate fine-tuning of GPT-3 Davinci due to budget. Overall, fine-tuning significantly improved accuracy for all models, with GPT-3 models uniformly improving by about 13 percentage points, and BERT/RoBERTa improving by about 25 points. Our best model after fine-tuning is RoBERTa, which is within 5\% of our human performance.

\begin{figure}[!ht]
    \centering
    \includegraphics[width=0.9\columnwidth]{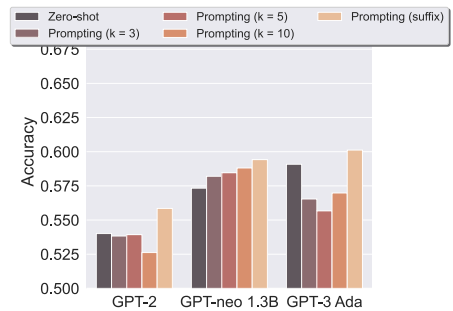}
    \caption{Comparison of prompting methods with autoregressive models}
    \label{prompting-graph}
\end{figure}

\paragraph{Prompting}
We also experiment with prompting methods. First, we use a simple \textit{suffix} prompting method, where we simply append the phrase "that is to say" between the metaphor and the interpretation, which we hypothesized may "explain" to the LM that the previous statement is figurative. We also evaluate the effectiveness of the \textit{examples} method, by appending $k$ random correct metaphor/interpretation pairs before the actual pair we are testing. The results of these experiments can be seen in \autoref{prompting-graph}. We found that the suffix method provided a small (1-2\%) improvement over the baseline, while the example method was generally ineffective.

\paragraph{Backward accuracies}

Note that the accuracies reported in this section are for the forward direction, and the backward direction is reported in \autoref{sec:backwards}. Backward accuracy is lower, with GPT-3 Curie for example having a ~7\% reduction in accuracy in the zero-shot case. This suggests that selecting a metaphorical expression to match a literal phrase is more challenging than the reverse for LMs. 

\paragraph{Paired Evaluation} Because our dataset is formatted as a Winograd schema, we can take advantage of \textit{group scoring} to evaluate models more stringently \cite{paired-acc}. We found that performance for autoregressive models plummeted under this evaluation scheme, while masked language models also suffered in accuracy. Human scores were the least affected. Details are in \autoref{sec:paired-acc}. This is likely related to the phenomenon found in \autoref{sec:same-preds}.

\subsection{Generation Results}
\label{sec:generation-res}

Next, we examine \textbf{if models can generate sensible interpretations for metaphors}.
Given the difficulty of evaluating text generation, compounded by the difficulty of figurative language, we opted for manual evaluation of one tenth of the test dataset 
using generations of the strongest auto-regressive model: GPT-3 Davinci ($\approx$175B parameters).

The metaphor was given as input to the model, and 4 completions were generated for each metaphor, with a maximum length of 100 tokens. Completions were also truncated to the first sentence, as initial experiments showed contradictory statements (e.g. "he was talented. But he was not very talented") were often generated across subsequent sentences. Suffix prompting was also used because of the lack of context, with "That is to say, " appended to each metaphor. Only the first sentence of the output was evaluated. The temperature parameter was determined through grid search through values [0.2, 0.4, 0.6, 0.8, 1] on a small separate set of metaphors. A human annotator inspected the generated completions and found that a temperature of 0.4 produced the most correct results. 

Three authors of the article labeled the completions generated by GPT-3 Davinci as correct, incorrect, or literal. In some cases, there were valid interpretations that were not the same as the answer given by crowdworkers, which were also marked correct. If the model simply restated the metaphor with no interpretation, the completion was marked as literal. Because some metaphors are ambiguous when presented without context, those examples were not counted.
Inter-rater reliability was moderate due to differing standards for correctness (Krippendorff's $\alpha$ = 0.5567). The majority vote was taken between annotators' judgments. 

GPT-3 Davinci's accuracy, counting literalized metaphors as incorrect, was 50.8\%. Not counting literalized metaphors, the accuracy was 63.9\%. In 37.7\% of cases, GPT-3 generated contradictory completions among the 4 completions. There was at least one correct completion for 78.1\% of the phrases, but only 19.3\% of phrases had all completions correct.
Examples of annotated generations can be found in \autoref{sec:generation}.

\section{Performance and Error Analysis}
\label{sec:error-analysis}

With these results in mind, we examine \textbf{what kinds of errors models make, and what factors make the task difficult.}. This is covered in \autoref{sec:error-analysis}. We find that autoregressive models rely on the predicted probability of each answer by itself to predict the answer, and that this holds true for all models, before and after training. We find that models have difficulty in interpreting "sarcastic" metaphors and sometimes inexplicably interpret very simple metaphors wrong. We also examine error typology according to the commonsense typology of \autoref{sec:metaphor-typology} and find that models improve significantly on object, visual, and social commonsense when trained, but not on cultural commonsense.

\subsection{Reliance on Probability of Answers}
\label{sec:same-preds}
\begin{figure}[!ht]
    \centering
    \includegraphics[width=0.8\columnwidth]{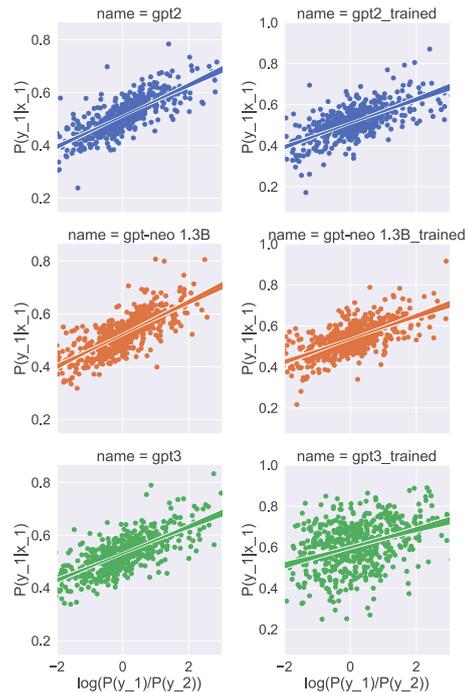}
    \caption{Models over-rely on predicted probability of the answer by itself to do their predictions. $y$-axis is  predicted probability of the first interpretation (answer) given metaphor while $x$-axis is log odds of the first interpretation. }
    \label{log-prob-pred}
\end{figure}

We find that models often rely solely on the predicted probability of answers $y_1$ and $y_2$ to make their final predictions, regardless of the context. This led the models to make the same prediction for the paired sentences in many cases. \autoref{log-prob-pred} and \autoref{tab:r-and-p-values} show that this trend improves with fine-tuning, and that GPT-3 is best able to disentangle the probability of $y_i$ and the probability of $P(y_i|x_i)$, but all three models show a heavy tendency to predict based on the relative probability of an answer alone. 

We hypothesize that this may be one reason why BERT and RoBERTa achieve the best finetuned performance; they use a contrastive finetuning strategy providing both the correct and incorrect options as input to the model. On the other hand, the GPT models were finetuned with only the correct option, making the comparison unfair. One way to finetune GPT models contrastively is to include both options into a cleverly engineered prompt, but we leave this as a direction for future work.


\begin{table}[!t]
    \centering
    \small
    \begin{tabular}{ccc}
        \toprule
         Model & $r$ & p \\ \toprule
         \multicolumn{3}{c}{Untrained} \\ \midrule
         GPT-2 & 0.8128 & $6.700 \times 10^{-136}$ \\
         GPT-neo & 0.7891 & $6.075\times10^{-123}$ \\
         GPT-3 & 0.7392 & $4.329\times10^{-100}$ \\ 
         \hline \multicolumn{3}{c}{Trained} \\ \midrule
         GPT-2 & 0.6765 & $6.700 \times 10^{-78}$\\
         GPT-neo & 0.6689 & $1.456\times10^{-75}$ \\
         GPT-3 & 0.4157 & $2.598\times10^{-25}$ \\
         \bottomrule
    \end{tabular}
    \caption{Spearman $r$-values and p-values between $P(y_i | x_i)$ and $P(y_i)$}
    \label{tab:r-and-p-values}
\end{table}

\subsection{Other Factors Influencing Correctness}

We also examined the influence of several other factors on correctness. The point-biserial correlation between length of the context phrase and the binary correctness value was -0.1544 with a p-value of $1.50\times10^{-7}$, indicating that longer phrases are harder to interpret correctly. The point-biserial correlation between answer probability and binary correctness was 0.1840, with a p-value of $3.50\times10^{-10}$, indicating that examples where the answer was already more probable were more likely to be answered correctly, in line with our findings that models tended to predict the answer that was already more plausible alone.

Furthermore, we conducted an analysis on subjects, objects, and relations as defined in \autoref{sec:sro}. We examined accuracy by part of speech patterns in each part of the metaphor, as well as by wordnet hypernyms present in each part of the metaphor. This is detailed in \autoref{sec:pos-acc} and \autoref{sec:pos-hypernym} \cite{wordnet}. We used NLTK for POS tagging \cite{nltk}.

\subsection{Qualitative Analysis of Error Trends}


\paragraph{Common Sense Knowledge}
We first examine the error tendencies by the type of common sense knowledge described in \autoref{sec:metaphor-typology}. \autoref{tab:commonsense-perf} summarizes accuracies for these types of commonsense questions compared to humans.

\begin{table}[!ht]
    \centering
    \small
    \begin{tabular}{ccccc}
        \toprule
        Model & Obj & Vis & Soc & Cul \\
        \toprule
        \multicolumn{5}{c}{Untrained} \\ \midrule
        GPT-2 & 52.17 & 52.07 & 55.38 & \textbf{58.42} \\
        GPT-neo & 56.38 & 55.62 & 56.01 & \textbf{62.10} \\
        GPT-3 Curie & 75.00 & 71.00 & 72.47 & \textbf{78.42} \\
        \midrule
        \multicolumn{5}{c}{Trained} \\ \midrule
        GPT-2 & 53.57 & 51.48 & \textbf{57.91} & 57.37 \\
        GPT-neo & 70.15 & \textbf{72.78} & 68.67 & 70.00 \\
        GPT-3 Curie & \textbf{87.50} & 84.62 & 83.86 & 83.16 \\
        BERT & 87.37 & \textbf{92.31} & 84.18 & 77.37 \\
        RoBERTa & 91.20 & \textbf{94.08} & 89.56 & 83.68 \\
        \midrule
        \midrule
        Human & 95.41 & \textbf{96.45} & 93.99 & 90.00\\
        \bottomrule
    \end{tabular}
    \caption{The performance of models across different commonsense categories, in terms of accuracy on examples annotated with that category (\%). The strongest category of each model is highlighted.}
    \label{tab:commonsense-perf}
\end{table}

We find that both humans and trained models tend to find object commonsense and visual commonsense metaphors easier to interpret. We find that as models improve, most of the performance gain comes from the object, visual, and social commonsense categories. Interestingly, the untrained models do quite well on cultural examples, but do not improve much on the culture category when trained. This makes sense, as cultural examples tend to be quite disparate, so training would not help as much with other examples.

\paragraph{Sarcastic Metaphors}
For both humans and LMs, many of the errors are "sarcastic" metaphors, such as saying "the girl was as bubbly as still water" to mean "the girl was bland", rather than "the girl was vivacious". These sentences can be difficult if the model or the human focuses on simple word association (bubbly with vivacious) without reading the entire sentence to understand the sarcasm. 

\paragraph{Inexplicable Errors}
We examined the errors made by GPT-3 Curie (trained) and found that there was little overlap with the errors made by humans.
Of the 64 human errors, 13 were also errors made by GPT-3. GPT-3 made many more "obvious" errors, such as predicting "The ball is a big red sun" to mean "the ball is small" rather than "the ball is big and red".
This is in contrast to sentences in which humans made errors, which often contained rare vocabulary or unfamiliar cultural references.

\section{Related work}
\label{sec:related}

\subsection{Figurative Language Identification}
\label{sec:related:identification}

Most existing work focuses on identifying figurative language at the word level. The VU Amsterdam Metaphor Corpus (VUA) is the largest available corpus of metaphorical language, annotated by humans \cite{vua}. Two shared tasks on metaphor identification have been run \cite{leong-etal-2018-report, leong-etal-2020-report}. Both have utilized the VUA corpus, and the latter also introduced the TOEFL corpus, sampled from essays written by non-native English speakers \cite{leong-etal-2020-report, beigman-klebanov-etal-2018-corpus}. Most participants in the shared tasks used neural models, notably BERT, RoBERTa, and Bi-LSTMs \cite{leong-etal-2020-report, bizzoni-ghanimifard-2018-bigrams, gao-etal-2018-neural, pramanick-etal-2018-lstm}. These models are generally improved when augmented with semantic data, such as concreteness and multimodal information.

Another line of work focuses on probing models to determine the extent of metaphor recognition. For example, BERT assigns higher pseudo-log-likelihood scores to metaphors than nonsense expressions, and its contextualized representations show some signs of contextualizing the object domain \cite{pedinotti-etal-2021-howling}. Another study uses linear probes trained on BERT layers to predict whether a word usage is literal or nonliteral, and finds that this can be done effectively, especially using middle layers as a representation \cite{probing-metaphors}, 

Despite the utility of these tasks and datasets, they have drawbacks. Most of the metaphor use is conventional, so this task does not capture novel metaphors well. The word-level annotation also does not lend itself well to capturing extended conceptual metaphors. Finally, metaphor interpretation may be a more difficult, although less studied, task.

\subsection{Figurative Language Interpretation}
\label{sec:related:interpretation}

Recent studies mostly focus on metaphor paraphrases, either through identification  \cite{bizzoni-lappin-2018-predicting} or generation \cite{shutova-2010-automatic, SU2017300, mao-etal-2018-word}.
However, there has not been as much work done on interpretation as on detection, and framing metaphor interpretation as a paraphrase task may not capture the emergent meaning of metaphors, such as the intended emotion, or the interaction of subject, relation and object in the metaphor \citep{tong2021, mohammad-etal-2016-metaphor}. 

Other work has focused on interpreting figurative language in narratives in context, based on plausible continuations of figurative language such as idioms and similes from stories \citep{Chakrabarty2021ItsNR} or dialogues \citep{jhamtani-etal-2021-investigating}. This represents a promising direction, and our work focuses on expanding our understanding of LMs' ability to interpret non-conventionalized metaphors. 

\subsection{Other Figurative Language Datasets}

We note that there are several other challenging NLI datasets available that contain figurative language, including the DNC corpus and the RTE dataset \cite{poliak-etal-2018-collecting, chakrabarty-etal-2021-figurative}. Other datasets, such as RiddleSense, explicitly test models through difficult commonsense inference, involving figurative language \cite{riddle}.

Our work is distinguished by the Winograd schema format, as this format provides a better guard against the possibility that models have simply memorized common word associations that occur in figurative language. Additionally, we specifically instructed crowdworkers to be creative, and this resulted in longer figurative phrases which require more detailed commonsense knowledge. It is likely that a fair number of these figurative phrases have never appeared in any training corpus. However, our figurative phrases also differ from riddles, as they are not supposed to be difficult to reason about, given that the source, relation, and object are properly contextualized.

\subsection{Human Language Processing}

Humans typically do not have any more difficulty processing metaphorical statements in context compared to literal statements \cite{fussel2008, metaphor-psych}. This may be because certain words serve as a \textit{dual reference}, that is to say, they simultaneously refer to a physical referent and an abstract superordinate category \cite{metaphor-psych}. For instance, "shark" may refer to literal sharks as well as anything that is considered vicious, leading to utterances such as "that lawyer is a shark". 

Metaphorical language processing has also been studied in second-language learners, in the case of idioms. In most cases, the meaning of an unfamiliar idiom is inferred from the context or from word association \cite{Cooper1999ProcessingOI, carston_wearing_2011, Wolff2000EvidenceFR}.

As LMs excel at word-association based tasks, this is an encouraging finding. However, there is still a gap between LM and human performance even in our task, in which one answer is obviously wrong when the input is correctly understood. 

We take into account that these results are for conventionalized figurative language and that some of the more creative phrases in this dataset may take a longer time to process for humans as well. This is especially true for non-native English speakers. However, the high human accuracy on this task with half the participants being non-native English speakers suggests that this was not a major barrier. 

\section{Conclusion}

We present a Winograd-like benchmark task to test the ability of LMs to reason about figurative language, based on a large-scale collection of creative metaphors written by humans. We find a large gap between LM zero-shot and human performance on this dataset, but show that models can be fine-tuned to perform well on this particular task. 

We hope that this work will encourage further study of nonliteral reasoning in LMs, especially in few-shot settings. Given that metaphorical reasoning may play a role in problem solving and linguistic creativity, the development of models, training methods, or datasets that enable metaphorical reasoning may improve models' abilities to reason creatively and draw analogies between situations that may appear to be different on the surface. One avenue we hope to investigate is multimodal metaphors, as this dataset currently includes only text-based metaphors. Nonliteral expressions also remain understudied cross-linguistically, but further work on identifying and interpreting metaphors in other languages may also improve the abilities of multilingual models.

\section{Ethical Considerations}
\subsection{Potential Risks}
Figurative language has the potential to be used in a harmful way, especially against minority and historically disadvantaged groups. Such language is often emotionally charged or used to insult others, so we took care to remove any examples that were potentially offensive, especially toward protected groups. We acknowledge that this was based on our own judgment and that generically insulting language (for instance, a metaphor that implies that someone is ugly) was not removed because it was not insulting toward any particular individual.

All examples from \datasetname are also in English, as it is the language that all authors speak, and this was a preliminary dataset, being the first of its type that the authors have worked on. However, figurative language is not just important in English, and we leave investigation of figurative language in other languages as future work.
\subsection{Terms of Use of Artefacts Used}

Additional datasets we used were the Winogrande dataset, SNLI, MNLI, FEVER-NLI and ANLI. Winogrande is licensed under the Apache 2.0 license, which allows modification and distribution, fitting our use case. SNLI is licensed under a Creative Commons Attribution ShareAlike 4.0 International license, which allows us to share and adapt the work as long as we give attribution. Most of MNLI is licensed under OANC, which allows free use. The fiction section of this dataset consists mostly of works in the public domain, but several stories are licensed: \textit{Seven Swords} is available under
a Creative Commons Share-Alike 3.0 Unported
License, while \textit{Living History} and \textit{Password Incorrect} are
available under Creative Commons Attribution
3.0 Unported Licenses. These licenses allow sharing and adaptation with attribution.
FEVER-NLI is licensed under an MIT license, which also allows modification, distribution, and reuse. ANLI is licensed under Creative Commons Attribution-NonCommercial 4.0 International, which also allows sharing and reuse as long as we give attribution.

Models used were GPT-2, GPT-neo, GPT-3, BERT and RoBERTa. GPT-2 and GPT-neo are licensed under an MIT license, which does not place any restrictions on their use. BERT is licensed under an Apache License 2.0, which allows modification and distribution. RoBERTa is licensed under the GNU General Public License v2.0. This fits our use case, as we are only running and studying the model. GPT-3 is licensed by Microsoft, and we used the public API to receive output.

\subsection{Computational Infrastructure and Computing Budget}

To run our computational experiments, we had access to a compute cluster, but minimal compute is needed to run the experiments in this paper. We generally did not use more than 2 GPUs at a time. The only models that required GPU parallelism were the GPT-neo models. An estimated 20 GPU hours are required.

Our computing budget was roughly 100 USD. We also used roughly 20 USD on credits for the GPT-3 API.

\section*{Acknowledgements}
We thank Pengfei Liu, Lyuyang Hu, and Chih-Hao Wang for helping us set up the leaderboard for this dataset on \href{https://explainaboard.inspiredco.ai/leaderboards?dataset=fig_qa}{Explainaboard}. We also thank Pengfei Liu for helping run GPT-3, Danish Pruthi for guidance on setting up the MTurk task, and all participants who contributed to the human benchmark. Lastly, we thank all the Turkers who contributed metaphors to the dataset.

This work was supported in part by a CMU Presidential Fellowship and National Science Foundation Award No. 1761548.

\bibliography{anthology,custom}
\bibliographystyle{acl_natbib}

\appendix

\section{Crowdsourcing Details}
\label{sec:crowdsourcing}

We crowdsource metaphorical expressions and their interpretations through Amazon Mechanical Turk. Workers were recruited from the United States and were limited to those who had a $> 98\% $ approval rating on the platform, and who had also completed more than 1000 Human Intelligence Tasks (HITs). Data collection was split into two stages: in the first stage, 1458 train examples, and all the dev and test examples were collected. In the second stage, the remaining 6558 training examples were collected. We identified some workers who created especially good examples in the first stage, and recruited them back for more examples in the second stage. Workers were paid \$0.33 for each pair of sentences and were asked to generate 3 pairs at a time. An author of this paper wrote an initial pilot set of sentences, and timed themselves while writing some sentences. They found that each pair took around 1 minute to write, though this varied (less creative examples took less time, while more creative examples took more time). This extrapolates to an hourly rate of $19.80$ USD, which is above the minimum wage in all US states, where workers were located.

Our HIT task was structured as follows: At the top of the page, the workers are shown the following instructions: "Your task is to generate three pairs of sentences with opposite or very different meanings, both of which contain rare/creative metaphors, which means metaphors that would not appear often in text on the Internet, books, social media, or news sites, but that can still be easily understood by people. For each metaphor, you should also provide a literal (non-metaphorical) sentence with the same meaning." Then, we display one example of a valid sentence pair. There is a button that opens a modal with more detailed instructions and some more valid/invalid examples for reference. Below that, we display three random words, which workers are encouraged to use in their sentences if they get stuck. Finally, we display three sets of 5 text fields for workers to fill in: one for the start phrase, two for each metaphorical phrase, and two for each literal interpretation. As the user types in each start phrase, we prepend a copy of their phrase before the corresponding metaphor fields in the UI using some embedded JavaScript, which we found helped reduce confusion and resulted in less improperly formatted responses. 

We launched many batches of these HITs until we had collected the desired quantity of data. Then, we converted the form responses into sentence pairs and validated each pair by hand before adding it to our dataset.

\section{Invalid Examples}
\label{sec:invalid}

Figurative language examples collected from crowdworkers were excluded if they (a) did not make sense given the meaning and metaphorical expression, (b) had grammar or spelling errors that rendered them unintelligible, or (c) did not follow the format specified by the task template. 

Examples are given below:

\begin{enumerate}
    \item Do not make sense given the meaning and the metaphorical expression
    \begin{table}[!ht]
        \centering
        \resizebox{\columnwidth}{!}{
        \begin{tabular}{cc}
            \toprule
            Paired sentences & Possible answers\\
            \toprule
            He was resourceful like toilet paper & He was very resourceful. \\
            He was resourceful like a mess & He wasn't resourceful at all \\
            \midrule
            The night was as long as a spool of thread & The night is long \\
            The night was as long as a winding road & The night dragged on \\
            \midrule
            the concert of the lession is a main and a major & we concert everyone\\
            the concert of the lession features & we concert our loved one \\
            \bottomrule
        \end{tabular}}
        \caption{Examples that were rejected due to being nonsensical.}
        \label{tab:invalid-nonsense}
    \end{table}
    \item Significant grammar or spelling errors
    \begin{table}[!ht]
        \centering
        \resizebox{\columnwidth}{!}{
        \begin{tabular}{cc}
            \toprule
            Paired sentences & Possible answers\\
            \toprule
            fallten data are very much trusted & fallten are nice \\
            fallten data are very valuable & flatten are safe \\
            \midrule
            CAR IS BIRD FEATHEAR & CAR SITE IS ROUGH\\
            CAR IS COTTON & CAR SITE IS HARD \\
            \midrule
            Inflation is as natural as Minnesota rainfall in June & Inflation is perfectly natural \\
            Inflation is as natural as Minnesota snowfall in June & Patient is in a natural result of other things \\
            \bottomrule
        \end{tabular}}
        \caption{Examples that were rejected due to having significant spelling or grammar errors.}
        \label{tab:invalid-spelling}
    \end{table}
    \item Do not follow format
    \begin{table}[!ht]
        \centering
        \resizebox{\columnwidth}{!}{
        \begin{tabular}{cc}
            \toprule
            Paired sentences & Possible answers\\
            \toprule
            \midrule
            This attack is as weak as a feather & The attack is useless \\
            This attack is as weak as a breeze & The attack doesn't work\\
            \midrule
            My car motor is dusty like old cave & Car motor is very rusty\\
            My car motor is dusty like abandon building & car motor is very dusty \\
            \midrule
            the writer is stuck between a rock And another hard place & He is just stuck doesnt have a choice \\
            the writer is stuck between a rock And a pebble & The writer can get over the pebble \\
            \bottomrule
        \end{tabular}}
        \caption{Examples that were rejected due to not following the specified format.}
        \label{tab:invalid-format}
    \end{table}
\end{enumerate}

Efforts were made to ensure that the final dataset contains no offensive content or personally identifiable information. WorkerID and other potentially personally identifying information were not included.

\section{Backward accuracies}
\label{sec:backwards}
\begin{table}[!ht]
    \centering
    \begin{tabular}{lcc}
        \toprule
        Model & Zero-shot & Fine-tuned (L) \\ \toprule
        GPT-2           & 52.18             & 52.00  \\
        GPT-neo 1.3B    & 54.36             & 63.44 \\
        GPT-3 Curie     & \textbf{58.46}             & \textbf{74.83}  \\
        \bottomrule
    \end{tabular}
    \caption{Zero-shot and finetuned backward auto-regressive model accuracies on the test set}
    \label{results-backwards}
\end{table}
\FloatBarrier

\FloatBarrier
\section{Paired accuracies}
\label{sec:paired-acc}
\begin{table}[!ht]
    \centering
    \begin{tabular}{cc}
    \toprule
        Model & \shortstack{Accuracy \\ (pairs correct)}\\
    \toprule
        GPT-2 zero-shot & 6.63 \\
        GPT-2 finetuned & 5.06 \\
    \midrule
        GPT-neo zero-shot & 10.3 \\
        GPT-neo finetuned & 10.3 \\
    \midrule
        GPT-3 Curie zero-shot & 17.4 \\
        GPT-3 Curie finetuned & 50.0 \\
    \midrule
        BERT finetuned & 70.6 \\
    \midrule
        RoBERTa finetuned & 80.4 \\
    \midrule
        Human & \textbf{89.7}
    \end{tabular}
    \caption{Accuracy for models on the test set, counted in terms of pairs of sentences in which both are correct (\%). Results are from one run.}
    \label{tab:paired-acc}
\end{table}
\FloatBarrier

\FloatBarrier
\section{Accuracy breakdown by Part-of-Speech}
\label{sec:pos-acc}
\FloatBarrier
\subsection{Subject}
\begin{table}[!ht]
    \centering
    \begin{tabular}{ccc}
        \toprule
        Part of speech & Accuracy & Frequency  \\ \toprule
        NN & 0.8569 & 538 \\
        PRP & 0.8526 & 156 \\
        PRP\$ NN & 0.9 & 110\\
        NN NN & 0.8889 & 63 \\
        DT NN & 0.8182 & 44 \\
        NN NN NN & 0.9375 & 32 \\
        JJ NN & 0.9167 & 12 \\
        \bottomrule
    \end{tabular}
    \caption{Accuracy breakdown and frequency of parts of speech in metaphor subjects. Only part-of-speech patterns with greater than 10 occurrences are shown.}
    \label{tab:accuracy-sub}
\end{table}
\FloatBarrier

\FloatBarrier
\subsection{Relation}
\begin{table}[!ht]
    \centering
    \begin{tabular}{ccc}
        \toprule
        Part of speech & Accuracy & Frequency  \\ \toprule
        VBZ NN IN & 0.8421 & 152 \\
        VBD RB JJ IN & 0.8904 & 146 \\
        VBZ RB JJ IN & 0.8889 & 99 \\
        VBZ & 0.8352 & 91 \\
        VBD NN IN & 0.8806 & 67 \\
        VBD & 0.9180 & 61 \\
        VBN IN & 0.9545 & 22 \\
        NN IN & 0.8636 & 22 \\
        VBD JJ IN & 0.9048 & 21 \\
        NNS IN & 0.8889 & 18 \\
        VBD IN & 0.8462 & 13 \\
        VBZ IN & 1.0 & 13 \\
        VBD RB VBN IN & 0.8182 & 11 \\
        \bottomrule
    \end{tabular}
    \caption{Accuracy breakdown and frequency of parts of speech in metaphor relations. Only part-of-speech patterns with greater than 10 occurrences are shown.}
    \label{tab:accuracy-rel}
\end{table}
\FloatBarrier

\FloatBarrier
\subsection{Object}
\begin{table}[!ht]
    \centering
    \begin{tabular}{ccc}
        \toprule
        Part of speech & Accuracy & Frequency \\ \toprule
        NN & 0.8788 & 429 \\
        NN NN & 0.8992 & 129 \\
        JJ NN & 0.8352 & 91 \\
        NN IN NN & 0.8372 & 43 \\
       JJ NN NN & 0.8710 & 31 \\
       NN NN NN & 0.9130 & 23 \\
       VBG NN & 0.9545 & 22 \\
       NN IN JJ NN & 0.6154 & 13 \\
       PRP\$ NN & 1.0 & 11\\
       JJ & 0.6364 & 11 \\
       NN IN NN NN & 0.8182 & 11 \\
       \bottomrule
    \end{tabular}
    \caption{Accuracy breakdown and frequency of parts of speech in metaphor objects. Only part-of-speech patterns with greater than 10 occurrences are shown.}
    \label{tab:accuracy-obj}
\end{table}
\FloatBarrier

\FloatBarrier
\section{Accuracy breakdown by hypernyms}
\label{sec:pos-hypernym}
\subsection{Subject}
\begin{table}[!ht]
    \centering
    \begin{tabular}{ccc}
        \toprule
        Synset & Accuracy & Frequency \\ \toprule
        adult.n.01 & 0.8736 & 182 \\
        male.n.02 & 0.8684 & 152 \\
        woman.n.01 & 0.7391 & 46 \\
        female.n.02 & 0.9130 & 46 \\
        show.n.03 & 0.875 & 24 \\
        product.n.02 & 0.8636 & 22 \\
        motor\_vehicle.n.01 & 0.9048 & 21 \\
        activity.n.01 & 0.8421 & 19 \\
        emotion.n.01 & 0.6667 & 18 \\
        publication.n.01 & 0.8333 & 18 \\
        feline.n.01 & 0.9375 & 16\\
        being.n.01 & 0.7143 & 14 \\
        performer.n.01 & 0.8333 & 12\\
        canine.n.02 & & 12\\
        body\_covering.n.01 & 0.8333 & 12 \\
        vessel.n.03 & 0.8333 & 12 \\
        sound.n.01 & 1.0 & 12 \\
        domestic\_animal.n.01 & 0.9167 & 12\\
        person.n.01 & 0.8 & 10 \\
        scheme.n.01 & 0.9 & 10 \\
        contestant.n.01 & 1.0 & 10 \\
        \bottomrule
    \end{tabular}
    \caption{Accuracy breakdown and frequency of WordNet hypernyms in metaphor subjects. Only hypernyms with 10 or greater occurrences are shown.}
    \label{tab:hypernym-sub}
\end{table}
\FloatBarrier

\FloatBarrier
\subsection{Object}
\begin{table}[!t]
    \centering
    \begin{tabular}{ccc}
        \toprule
        Synset & Accuracy & Frequency \\ \toprule
        time\_period.n.01 & 0.85 & 20 \\
        natural\_object.n.01 & 0.8947 & 19 \\
        person.n.01 & 0.8824 & 17 \\
        large\_integer.n.01 & 0.9286 & 14 \\
        adult.n.01 & 1.0 & 14 \\
        solid.n.01 & 0.9167 & 13 \\
        male.n.02 & 1.0 & 13\\
        child.n.02 & 0.8333 & 12 \\
        body\_of\_water.n.01 & 0.75 & 12 \\
        body\_covering.n.01 & 0.8333 & 12 \\
        digit.n.01 & 0.9167 & 12 \\
        region.n.01 & 0.8182 & 11\\
        beverage.n.01 & 0.8182 & 11 \\
        juvenile.n.01 & 0.8182 & 11\\
        container.n.01 & 0.9 & 10 \\
        rodent.n.01 & 0.9 & 10\\
        feline.n.01 & 0.7 & 10 \\
        building.n.01 & 0.8 & 10\\
        time\_unit.n.01 & 0.9 & 10 \\
        travel.v.01 & 0.7 & 10 \\
        \bottomrule
    \end{tabular}
    \caption{Accuracy breakdown and frequency of WordNet hypernyms in metaphor objects. Only hypernyms with 10 or greater occurrences are shown.}
    \label{tab:hypernym-obj}
\end{table}
\FloatBarrier

\FloatBarrier
\section{Generation examples}
Generation examples can be found in \autoref{tab:generation-examples}.
\label{sec:generation}
\begin{table*}[!t]
    \centering
    \resizebox{\textwidth}{!}{
    \begin{tabular}{ccc}
        \toprule
        Startphrase & Completion & Label\\
        \toprule
        His temper was a marshmallow. That is to say, & he was very easily angered, and once he was, he was very hard to cool down & Incorrect \\ 
        \midrule
        He's got the swimming ability of a pack of dolphins. That is to say, & he's got none & Incorrect \\
        \midrule
        The villain is as beautiful as the Joker. That is to say, & he's not & Correct \\
        \midrule
        The child has the energy of a sloth. That is to say,& he is lazy & Correct \\
        \midrule
        The girl moved as fast as a turtle. That is to say, & she moved at a turtle's pace & Literal \\
        \midrule
        The sadness of her death was a flea.That is to say, & It was a flea that was a sadness & Literal \\
        \bottomrule
        
    \end{tabular}}
    \caption{Examples of completions generated by GPT-3 Davinci.}
    \label{tab:generation-examples}
\end{table*}
\end{document}